\documentclass[letterpaper, 10 pt, journal, twoside]{misc/ieeetran}

\IEEEoverridecommandlockouts

\usepackage{lipsum}

\usepackage{amsfonts,amssymb,amsmath,bm}

\usepackage{graphicx}
\usepackage{siunitx}
\usepackage{glossaries}

\usepackage{etoolbox}
\makeatletter
\patchcmd{\@makecaption}
{\scshape}
{}
{}
{}
\makeatother

\usepackage{color}
\usepackage{ifthen}
\usepackage{flushend}
\newboolean{authnotes}
\usepackage{hyperref}

\usepackage{tabularx}
\usepackage{multirow}
\usepackage{multicol}
\usepackage{makecell}
\usepackage{booktabs}

\usepackage{comment}

\makeatletter
\let\NAT@parse\undefined
\makeatother

\usepackage[numbers]{natbib}
\bibpunct{[}{]}{,}{n}{}{;}

\setboolean{authnotes}{true}

\ifthenelse{\boolean{authnotes}}
{
\newcommand{\todo}[1]{{{\bf\color{magenta} TODO: #1}}}

}
{
\newcommand{\todo}[1]{}
\newcommand{\nf}[1]{}
\newcommand{\db}[1]{}
\newcommand{\st}[1]{}
}

\newcommand{\R}{\mathbb{R}}

\def\1{\bm{1}}

\def\RR{\mathbb{R}}

\def\vtheta{{\bm{\theta}}}
\def\va{{\bm{a}}}

\def\vd{{\bm{d}}}
\def\ve{{\bm{e}}}

\def\vg{{\bm{g}}}

\def\vq{{\bm{q}}}
\def\vr{{\bm{r}}}
\def\vs{{\bm{s}}}

\def\vx{{\bm{x}}}

\def\vtheta{{\boldsymbol{\theta}}}
\def\vtau{{\boldsymbol{\tau}}}

\def\mD{{\bm{D}}}

\def\mI{{\bm{I}}}
\def\mJ{{\bm{J}}}
\def\mK{{\bm{K}}}

\def\mM{{\bm{M}}}

\def\mS{{\bm{S}}}

\DeclareMathAlphabet{\mathsfit}{\encodingdefault}{\sfdefault}{m}{sl}
\SetMathAlphabet{\mathsfit}{bold}{\encodingdefault}{\sfdefault}{bx}{n}
\newcommand{\tens}[1]{\bm{\mathsfit{#1}}}

\def\tF{{\tens{F}}}

\def\gX{{\mathcal{X}}}

\newcommand{\E}{\mathbb{E}}

\DeclareMathOperator*{\argmax}{arg\,max}

\makeatletter
\newcommand{\StatexIndent}[1][3]{\setlength\@tempdima{\algorithmicindent}\Statex\hskip\dimexpr#1\@tempdima\relax}
\makeatother 
\title{Benchmarking Structured Policies \\ and Policy Optimization for\\ Real-World Dexterous Object Manipulation}

\author{Niklas Funk$^{*1}$, Charles Schaff$^{*2}$, Rishabh Madan$^{*3}$, Takuma Yoneda$^{*2}$, Julen Urain De Jesus$^{1}$ \\
       Joe Watson$^{1}$, Ethan K. Gordon$^{4}$, Felix Widmaier$^{5}$, Stefan Bauer$^{5}$, Siddhartha S. Srinivasa$^{4}$\\
        Tapomayukh Bhattacharjee$^{3}$, Matthew R. Walter$^{2}$, Jan Peters$^{1}$\thanks{Manuscript received: April, 1, 2021; Accepted June, 18, 2021.}\thanks{This paper was recommended for publication by Editor T. Asfour upon evaluation of the Associate Editor and Reviewers' comments.}
\thanks{This work was supported by the European Union’s Horizon 2020 program under grant agreement No. 640554 (SKILLS4ROBOTS). Niklas Funk acknowledges the support of the Nexplore/HOCHTIEF Collaboration Lab at TU Darmstadt. Charles Schaff and Takuma Yoneda were supported in part by the National Science Foundation under Grant No.\ 1830660.
Research reported in this publication was supported by the Eunice Kennedy Shriver National Institute Of Child Health \& Human Development of the National Institutes of Health under Award Number F32HD101192. The content is solely the responsibility of the authors and does not necessarily represent the official views of the National Institutes of Health. This work was also (partially) funded by the National Science Foundation IIS (\#2007011), National Science Foundation DMS (\#1839371), the Office of Naval Research, US Army Research Laboratory CCDC, Amazon, and Honda Research Institute USA. (\textit{Corresponding author: Niklas Funk})}
\thanks{*Equal contribution. Names are displayed in a random order.}\thanks{$^{1}$Niklas Funk, Joe Watson, Julen Urain de Jesus and Jan Peters are with the Department of Computer Science, Technical University of Darmstadt, Darmstadt, Germany 
{\tt\small \{niklas,julen,joe,jan\}@robot-learning.de}}\thanks{$^{2}$Charles Schaff, Takuma Yoneda and Matthew R. Walter are with the Toyota Technological Institute at Chicago, Chicago IL, USA,
        \tt\small \{cbschaff, takuma, mwalter\}@ttic.edu}\thanks{$^{3}$Rishabh Madan and Tapomayukh Bhattacharjee are with the Department of     Computer Science, Cornell University,
        Ithaca, NY, USA,
        {\tt\small \{rm773, tapomayukh\}@cornell.edu}}\thanks{$^{4}$Ethan K. Gordon and Siddhartha S. Srinivasa are with the University of Washington,
        Seattle WA, USA,
        {\tt\small \{ekgordon, siddh\}@uw.edu}}\thanks{$^{5}$Stefan Bauer and Felix Widmaier are with the Max Planck Institute for Intelligent Systems, Tübingen, Germany,
        \tt\small \{felix.widmaier, stefan.bauer\}@tue.mpg.de}\thanks{Digital Object Identifier (DOI): 10.1109/LRA.2021.3129139}
}

\usepackage{fancyhdr}
 \newcommand{\mytitle}{\textbf{Accepted version.} To appear in \textit{IEEE Robotics and Automation Letters}.  DOI:
10.1109/LRA.2021.3129139\\
\copyright 2021 IEEE. Personal use of this material is permitted.
Permission from IEEE must be obtained for all other uses, in any current or future media, including reprinting/republishing this material for advertising or promotional purposes, creating new collective works, for resale or redistribution to servers or lists, or reuse of any copyrighted component of this work in other works.} 
\begin{document}

\markboth{IEEE Robotics and Automation Letters. Preprint Version. Accepted July, 2021}
{Funk \MakeLowercase{\textit{et al.}}: Benchmarking Structured Policies and Policy Optimization for Real-World Dexterous Object Manipulation}

\newacronym{bo}{BO}{bayesian optimization}
\newacronym{rpl}{RPL}{residual policy learning}
\newacronym{rrc}{RRC}{Real Robot Challenge}
\glsunset{rrc}

\maketitle

\thispagestyle{fancy}

\begin{abstract}
Dexterous manipulation is a challenging and important problem in robotics.
While data-driven methods are a promising approach, current benchmarks require simulation or extensive engineering support due to the sample inefficiency of popular methods.
We present benchmarks for the TriFinger system, an open-source robotic platform for dexterous manipulation and the focus of the 2020 Real Robot Challenge.
The benchmarked methods, which were successful in the challenge, can be generally described as \emph{structured policies}, as they combine elements of classical robotics and modern policy optimization. 
This inclusion of inductive biases facilitates sample efficiency, interpretability, reliability and high performance. 
The key aspects of this benchmarking is validation of the baselines across both simulation and the real system, thorough ablation study over the core features of each solution, and a retrospective analysis of the challenge as a manipulation benchmark.

\end{abstract}

\begin{IEEEkeywords}
Dexterous Manipulation, Grasping, Performance Evaluation and Benchmarking
\end{IEEEkeywords}

\section{INTRODUCTION}
\begin{figure}
    \centering
    \includegraphics[width=\columnwidth]{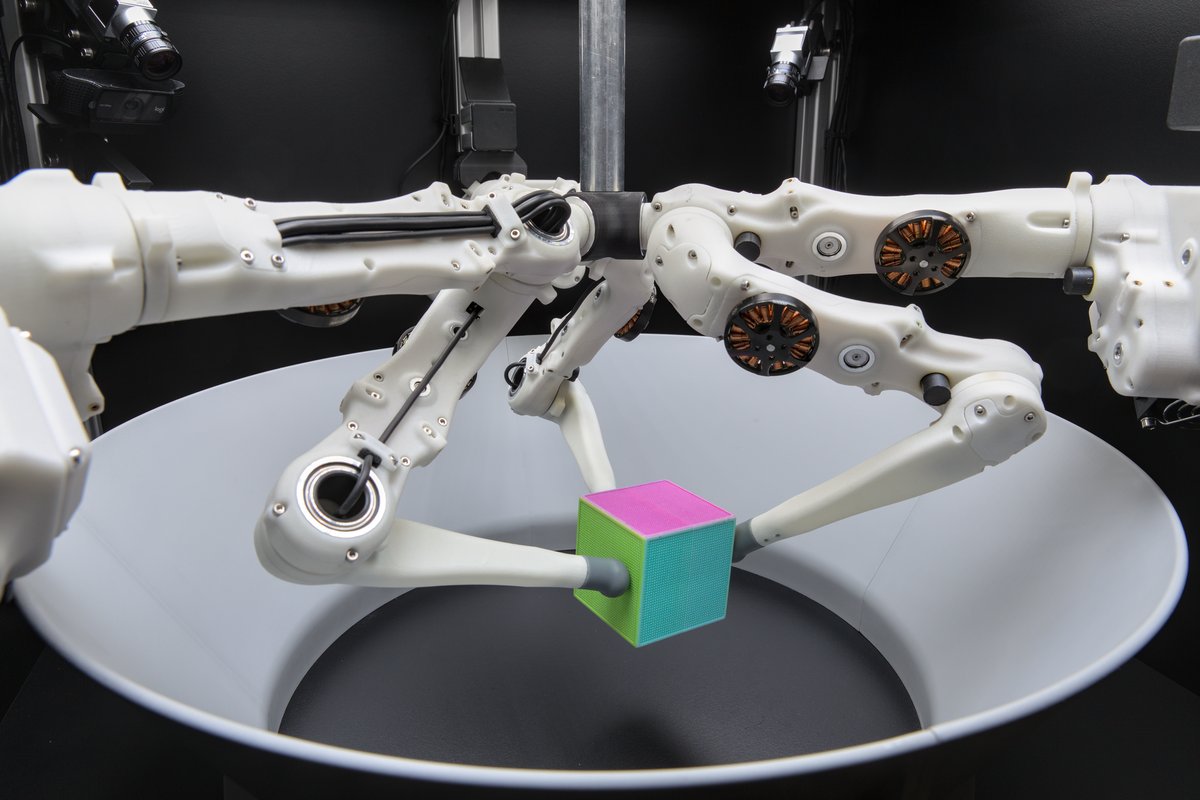}
    \caption{The TriFinger platform performing a task in the Real Robot Challenge, bringing a cube object to a desired pose.}
    \label{fig:RRCsystem}
    \vspace{-0.5cm}
\end{figure}

\IEEEPARstart{D}{exterous} manipulation is a challenging problem in robotics
that has impactful applications across industrial and domestic settings.
Manipulation is challenging due to a combination of environment interaction, high-dimensional control and required exteroception.
As a consequence, designing high-performance control algorithms for physical systems remains a challenge.
Due to the complexity of the problem, data-driven approaches to dexterous manipulation are a promising direction.
However, due to the high cost of collecting data with a physical manipulator 
and the sample efficiency of current methods, the robot learning community has primarily focused on simulated experiments and benchmarks~\cite{haarnoja2018soft, fujimoto18, popov2017data, charlesworth2020solving, Rajeswaran-RSS-18}. While there have been successes on hardware for various manipulation tasks~\cite{gu2017deep,openai2019solving,akkaya2019solving,PDDM}, the hardware and engineering cost of reproducing these experiments can be prohibitive to most researchers.

In this work, we investigate several approaches to dexterous manipulation using the TriFinger platform \cite{wuthrich2020trifinger},
an open-source manipulation robot.
This research was motivated by the `Real Robot Challenge' (RRC),\footnote{See \url{https://real-robot-challenge.com}.} where the community was tasked with designing manipulation agents on a farm of physical TriFinger systems.
A common theme among the successful solutions is their use of \emph{structured policies}, methods that combine elements of classical robotics and modern machine learning to achieve reliability, sample efficiency and high performance.
We summarize the solutions here and analyse their performance through ablation studies to understand which aspects are important for real-world manipulation and how these characteristics can be appropriately benchmarked.

The main contributions of our work are as follows. 
We introduce \textbf{three independent structured policies} for tri-finger object manipulation and \textbf{two data-driven optimization schemes}.
We perform a detailed \textbf{benchmarking and ablation study} across policy structures and optimization schemes, with evaluations both in simulation and on several TriFinger robots.
The paper is structured as follows: Section II discusses prior work, Section III introduces the TriFinger platform, Section IV describes the structured policies, Section V presents the approaches to structured policy optimization, Section VI details the experiments, and Section VII discusses the findings.
 
\section{RELATED WORK}

Much progress has been made for manipulation in structured environments with \textit{a priori} known object models through the use of task-specific methods and programmed motions. However, these approaches typically fail when the environment exhibits limited structure or is not known exactly. 
The contact-rich nature of manipulation tasks naturally demands compliance, which can be achieved through soft materials on the hardware level~\cite{soft_robots_Iida}.
The need for compliant motion has also led to the development of control objectives like force control~\cite{mason1981compliance}, hybrid position/force control~\cite{raibert1981hybrid}, and impedance control~\cite{hogan1985impedance}. 
Operational space control~\cite{khatib1987unified} has been monumental in developing compliant task space controllers that use torque control.

\textbf{Data-driven Manipulation} 
Given the complexity of object manipulation due to both the hardware and task, data-driven approaches are an attractive means to avoid the need to hand-design controllers, while also providing potential for improved generalizability and robustness.
For control, a common approach is to combine learned models with optimal control, such as guided policy search \cite{local_models} and model predictive control (MPC)~\cite{PDDM, POLO}.
Model-free reinforcement learning (RL) has also been applied to manipulation~\cite{8276248, HoofHumanoids2015}, including deep RL~\cite{Rajeswaran-RSS-18,openai2019solving}, which typically requires demonstrations or simulation-based training due to sample complexity. 
Data-driven methods can also improve grasp synthesis~\cite{6672028, veiga2012towards}.

\textbf{Structured Policies} Across machine learning, \emph{inductive biases} provide a means to introduce domain knowledge to improve sample efficiency, interpretability and reliability. 
In the context of control, inductive biases have been applied to enhance models for model-based RL \cite{Lutter_PIICRA_2021}, or to policies to simplify or improve policy search.
Popular structures include options \cite{Daniel2016ECML}, dynamic movement primitives~\cite{NIPS2002_23c97e9c, Paraschos_NIPS_2013a, bahl2020neural}, autoregressive models~\cite{autoregressive}, MPC~\cite{pmlr-v84-kamthe18a} and motion planners \cite{yamada2020mopa, xiali2020relmogen}.
Structure also applies to the action representation, as acting in operational space, joint space or pure torque control affects how much classical robotics can be incorporated into the complete control scheme \cite{martin2019iros}. 

\textbf{Residual Policy Learning}
\Gls{rpl}~\cite{Silver2018ResidualPL, johannink2019residual} provides a way to enhance a given base control policy by learning additive, corrective actions using RL~\cite{sutton18}.
This allows well developed tools in robotics---such as motion planning, PID controllers, etc.---to be used as an inductive bias for learned policies in an RL setting. This combination
improves sample efficiency and exploration, leading to policies that can outperform classical approaches and pure RL.

\textbf{Bayesian Optimization for Control}
\Gls{bo}~\cite{snoek2012practical} is a sample-efficient black-box optimization method that leverages the epistemic uncertainty of a Gaussian process model of the objective function to guide optimization.
It can be used for hyperparameter optimization, policy search, sim-to-real transfer and grasp selection~\cite{snoek2012practical, marco2016automatic, muratoreDR, veiga2012towards}.

\textbf{Benchmarking Manipulation}
Early work on benchmarking dexterous manipulation was mainly focused on simulation \cite{ulbrich2011opengrasp}. Lately, there has been an increased interest in real-world setups. Yet, most require large, expensive hardware and complex software solutions, including perception modules \cite{leitner2017acrv,cruciani2020benchmarking}. In contrast, this work builds on the \gls{rrc}, which provides remote access to TriFinger platforms, allowing for an exclusive focus on developing effective manipulation strategies. 
\section{TRI-FINGER OBJECT MANIPULATION} \label{sec:trifingerchallenge}

Fig.~\ref{fig:RRCsystem} shows the TriFinger robot \cite{wuthrich2020trifinger}.
The inexpensive, compact and open-source platform can be easily recreated and serves as the basis for the \gls{rrc}, which aims to promote state-of-the-art research in dexterous manipulation on real hardware.

\textbf{About the Robot}
The robot consists of three identical
``fingers" with three degrees of freedom each.
Its robust design, together with several safety measures, allow for running learning algorithms directly on the real robot, even if they
send unpredictable or random commands. 
The robot can be controlled with either torque or position commands at a rate of
1\,kHz.  It provides measurements of angles, velocities and
torques of all joints at the same rate.
A vision-based tracking system provides the pose of the manipulated object at a frequency of 10\,Hz. 
Users can interact with the platform by submitting experiments and downloading the logged data.

\textbf{The Real Robot Challenge}
\label{challenge}
The \emph{Real Robot Challenge 2020} 
involved three phases using the TriFinger robot: simulation (Phase 1), and the real robot manipulating a cube (Phase 2), and a cuboid (Phase 3).
While our methods were used in all three phases, we focus on Phase 2 because it involved a real robot and a simpler object that afforded much more robust and reliable pose estimation, which is crucial for a fair comparison of the approaches.
This phase tasks the robot with moving a cube (Fig.~\ref{fig:RRCsystem}) from the center of the arena to a desired goal pose.
The cube weighs about 94\,g, has a side-length of 65\,mm, a structured surface to facilitate grasping, and differently colored sides to help vision-based pose estimation. 
\begin{table}[!t]
    \vspace{5pt}
    \centering
    \begin{tabular}{lrrrrr}
        \toprule
        Strategy & Level 1 & Level 2 & Level 3 & Level 4 & Total \\
        \midrule
        MP-PG & -5472 & -2898 & -9080 & -21428 & -124221 \\
        CPC-TG & -3927 & -4144 & -4226 & -48572 & -219182 \\
        CIC-CG & -6278 & -13738 & -17927 & -49491 & -285500 \\
        \bottomrule
        \noalign{\vskip 2mm}  
    \end{tabular}
    \caption{Final results of the \gls{rrc} Phase 2.  The score of each level is the average reward over multiple runs. 
    \emph{Total} is a weighted sum of returns over the four levels.
    The strategies of our independent submissions to the challenge are: Grasp and Motion planning (MP-PG), Cartesian position control with Triangle Grasp (CPC-TG), and Cartesian impedance control with Center of Three Grasp (CIC-TG).
    They ranked 1st, 2nd and 4th in the competition.
    }
    \vspace{-.75cm}
    \label{tab:challenge_results_phase2}
\end{table}

The phase is subdivided into four difficulty levels (L1--L4).
We focus on the final two, which involve reaching a goal position (L3) and pose (L4) sampled from anywhere in the workspace.
Thus, for L3 the reward only reflects the position error of the cube and
is computed as a normalized weighted sum of error $\ve = (e_x, e_y, e_z)$ between the actual and goal positions:
$r_{3} = -\left(\frac{1}{2} \cdot  \frac{\left \| \ve_{xy} \right \|}{d_{xy}} + \frac{1}{2} \cdot
  \frac{|e_z|}{d_z} \right)$,
with $d_{xy}$ and $d_z$ the range on the x/y-plane and z-axis, respectively.
For L4, the orientation error is computed as the normalized magnitude of the rotation $\vq$ (given as quaternion) between actual and goal orientation $\text{err}_\text{rot} = 2\operatorname{atan2}(||\vq_{xyz}||, |q_w|)/\pi$. Thus, the reward is given by $r_4 = (r_{3} - \text{err}_\text{rot})/2$. 

This paper benchmarks the solutions of three independent submissions to the challenge. Table \ref{tab:challenge_results_phase2} shows their performance. 
\section{STRUCTURED POLICIES}
\begin{figure}
    \vspace{2.5pt}
    \centering
    \includegraphics[width=\columnwidth]{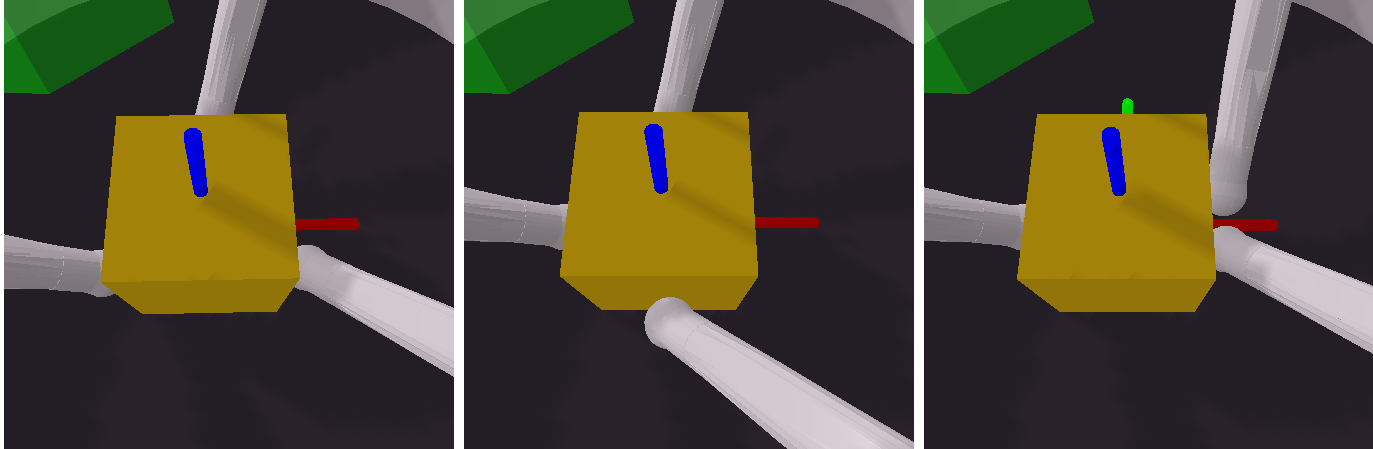}
    \caption{Depicting the different grasp strategies. From left to right: triangle grasp (TG), center of three grasp (CG), and opposite faces grasp (OG).
    All permutations of fingers and vertical faces for the CGs and OGs are considered for the planned grasp (PG) heuristic.}
    \label{fig:GraspStrategies}
    \vspace{-.5cm}
\end{figure}

This section describes the structured controllers considered as baselines.
The controllers share a similar high-level structure and can be broken into three main components: cube alignment, grasping, and movement to the goal pose.
We will discuss the individual grasp and goal approach strategies, and then briefly mention cube alignment.
For a visualization of each grasping strategy, see Fig.~\ref{fig:GraspStrategies}.
For a more detailed discussion of each controller and alignment strategy, please see the reports submitted for the RRC competition: motion planning~\cite{yoneda2021grasp}, Cartesian position control~\cite{madanrrc2020}, and Cartesian impedance control~\cite{anonymous2020modelbased}. 
The code is publicly available\footnote{\url{https://github.com/cbschaff/benchmark-rrc}} and demo videos are added in the supplementary material.

\subsection{Grasp and Motion Planning (MP)}
\label{gmp}
When attempting to manipulate an object to reach a desired goal pose, the grasp must be carefully selected such that it can be maintained throughout the manipulation.
Many grasps that are valid at the current object pose may fail when moving the object.
To avoid using such a grasp, 
we consider several heuristic grasp candidates and attempt to plan a path from the initial object pose to the goal pose under the constraint that
the grasp is maintained at all points along the path. 
Path planning involves first selecting a potential grasp and then running a rapidly exploring random tree (RRT)~\cite{lavalle1998rapidly} to generate a plan in task space, using the grasp to determine the fingers' joint positions. If planning is unsuccessful within a given time frame, we select another grasp and retry. We consider two sets of heuristic grasps, one with fingertips placed at the center of three vertical faces and the other with two fingers on one face and one on the opposite face. In the unlikely event that none of those heuristic grasps admits a path to the goal, we sample random force closure grasps~\cite{modrob} until a plan is found. Throughout this paper we refer to this method for determining a grasp as {\bf Planned Grasp (PG)}.

To move the object to the goal pose, this approach then simply executes the motion plan by following the waypoints for all fingers using a PD position controller without requiring any further cube pose estimates.
After execution, to account for errors such as slippage or an inaccurate final pose, we iteratively append waypoints to the plan in a straight line path from the current object position to the goal position.

\subsection{Cartesian Position Control (CPC)}
\label{pos_control}
Much of the tasks presented in the challenge can be solved using carefully designed position-based motion primitives. We build upon this intuition and implement a controller that uses Cartesian space position as the reference and outputs joint torques to actuate the robot.
To do so, we first reduce the TriFinger joint space $\vq \in \R^9$ to the 3D Cartesian position space of each end effector $\vx \in \R^{9}$, discarding finger orientation due to the rotational near-symmetry of the end effectors.
We retrieve the Jacobian matrix, $\mJ := \frac{\partial{\vx}}{\partial{\vq}}$. The Jacobian-inverse converts the task velocity to joint velocity according to $\dot{{\vq}} = {\mJ}^{-1}\dot{{\vx}}$. ${\mJ}$ may be singular or non-square, so we use its damped pseudo inverse to guarantee a good solution at the cost of slight bias: $\dot{{\vq}} = {\mJ}^\top\left({\mJ}{\mJ}^\top + \lambda{\mI}\right)^{-1}\dot{{\vx}}$. We combine this with gravity compensation torques to command gravity-independent linear forces at each finger tip.

We build upon these linear force commands to create position-based motion primitives. Given a target position for each fingertip, we construct a feedback controller with tuned PID gains~\cite{ziegler1942optimum} coupled with some minor adjustments in response to performance changes. This approach works well in simulation, but for the real robot, it results in the fingers getting stuck in intermediate positions in some cases. Based on the limited interaction afforded by remote access to the robots, this could be attributed to static friction causing the motors to stop. 
We use a simple gain scheduling mechanism that varies the gains exponentially (up to a clipping value) over a specified time interval, which helps to mitigate this degradation in performance by providing the extra force required to keep the motors in motion.

\textbf{Triangle Grasp (TG)}
The above controller is combined with a grasp that places fingers on three of the vertical faces of the cube. The fingers are placed such that they form an equilateral triangle~\cite{roa2015grasp}. This ensures that the object is in force closure and that the fingers can easily apply forces on the center of mass of the cube in any direction. 
\subsection{Cartesian Impedance Control (CIC)}
\label{imp_control}
Third, we present a Cartesian impedance controller (CIC) \cite{modrob,wimbock2012comparison, biagiotti2003cartesian, pfanne2020object}.
Using CIC enables natural adaptivity with the environment for object manipulation by specifying second-order impedance dynamics.
This avoids having to learn the grasping behaviour through extensive experience and eludes complex trajectory optimization that must incorporate contact forces and geometry.
Avoiding such complexity results in a controller that has adequate baseline performance on the real system that can then be further optimized. 

For the desired Cartesian position of the $i^\text{th}$ fingertip $\vx_i$, we define $\bar{\vx}_i$ to be the error between this tip position and a reference position inside the cube $\vx_r$, i.e., $\bar{\vx}_i = \vx_r - \vx_i$. 
We then define an impedance controller for $\bar{\vx}_i$, a second-order ODE that can be easily interpreted as a mass-spring-damper system with parameters ${\mM, \mD, \mK}$. 
The damping factor was zeroed for more robustness to the measurement noise that is present in the vision-based estimate of the cube's pose. Converting this Cartesian space control law back to joint coordinates results in ${\vtau_{1,i} = \mM(\vq) \mJ^{-1} \ddot{\bar{\vx}}_i}$, where $\vtau_{1,i}$ denotes the torques to be applied to finger $i$. 

To perform cube position control, we follow the ideas proposed by Pfanne et al.~\cite{pfanne2020object} and design a proportional control law that perturbs the center of cube $\vx_c$ based on the goal position $\vx_g$: $\hat{\vx}_r=\vx_r+ \mK_1 (\vx_g-\vx_c)$. 

Since the above components do not consider the fingers as a whole, they were limited in controlling the orientation of the cube.
Contact forces were also passively applied rather than explicitly considered.
To incorporate these additional considerations, we superimpose four torques.
These include the previously discussed position control and gravity compensation as well as three contact and rotational terms that we describe next, such that $\vtau_i = \textstyle\sum_{j=1}^{4} \vtau_{j,i}$.

To also allow one to directly specify the force applied by each finger, we introduce an additional component $\vtau_{2,i}=\mJ^\top \tF_{2,i}$, where $\tF_{2,i}$ is the force applied by finger $i$.
We chose $\tF_{2,i}$ to be in the direction of the surface normal of the face where finger $i$ touches the cube ($\tF_{2,i}=\mK_2 \vd_i$).
However, in order to not counteract the impedance controller, the resulting force of this component $\tF_{\text{res}}= \textstyle\sum_i \tF_{2,i}$ should be zero. 

By solving
${-\tF_{\text{res}} =[\mJ^{-\top}, \mJ^{-\top}, \mJ^{-\top}] [\vtau_{3,1}, \vtau_{3,2}, \vtau_{3,3}]^\top} $
for $\vtau_{3,i}$, this is ensured.
All previous components ensure a stable grasp closure. This is essential for the following orientation control law. Neglecting the exact shape of the cube, we model the moment that is exerted onto the cube as ${\Omega = \sum \vr_i \times \tF_{4,i} = \sum \mS_{r_i} \tF_{4,i}} $, where $\vr_i=-\bar{\vx}_i / {|{\bar{\vx}_i}|_2}$ denotes the vector pointing from the center of the cube towards the finger position, $\mS_{r_i}$ is the respective skew-symmetric matrix, and $\tF_{4,i}$ an additional force that should lead to the desired rotation. 
The goal is now to realize a moment proportional to the current rotation errors, which are provided in the form of an axis of rotation $\vr_{\phi}$ and its magnitude $\phi$. Thus, the control law yields $\Omega = \mK_3 \phi \vr_{\phi}$. 
We achieve $\Omega$ by solving
${ \Omega = [\mS_{\vr_1} \mJ^{-\top}, \mS_{\vr_2} \mJ^{-\top}, \mS_{\vr_3} \mJ^{-\top}] [\vtau_{4,1}, \vtau_{4,2}, \vtau_{4,3}]^\top}, $
for $\vtau_{4,i}$.

\textbf{Center of Three Grasp (CG)}
The above controller is combined with a grasp that places the fingers in the center of three of the four faces perpendicular to the ground plane. The three faces are selected based on the task and goal pose with respect to the current object pose. When only a goal position is specified (L1--L3) the face that is closest to the goal location is not assigned any finger, ensuring that the cube can be pushed to the target. 
For L4 cube pose control, 
two fingers are placed on opposite faces such that the line connecting them is close to the axis of rotation. To avoid colliding with the ground, the third finger is placed such that an upward movement will yield the desired rotation.

\subsection{Cube Alignment}

While we will focus our experiments on the above methods, another component was necessary for the teams to perform well in the competition. Moving an object to an arbitrary pose often requires multiple interactions with the object itself.
Therefore, teams had to perform some initial alignment of the cube with the goal pose. All three teams independently converged on a sequence of scripted motion primitives to achieve this. These primitives consisted of heuristic grasps and movements to (1) slide the cube to the center of the workspace, (2) perform a $90$ degree rotation to change the upward face of the cube, and (3) perform a yaw rotation. Following this sequence, the cube is grasped and moved to the goal pose. 

\section{POLICY OPTIMIZATION}
In addition to the approaches described above, the teams experimented with two different optimization schemes: Bayesian optimization and residual policy learning. In this section, we will briefly introduce these methods.

\textbf{Bayesian Optimization}
\gls{bo} is a black-box method to find the global optimum of an unknown function $f: \gX \xrightarrow[]{} \RR$. This is achieved by approximating $f$ with a nonparametric Gaussian process model $\mathcal{GP}(0,k)$.
Our Gaussian processes provide a zero-mean prior distribution over functions with explicit uncertainty estimates in which prior knowledge about $f$ is encoded through the covariance function $k:\gX\times\gX\rightarrow\RR$.
Using this prior, BO selects the next point $\vx \in \gX$ from a set of candidates such that $\vx$ maximizes some criterion called the acquisition function. $f(\vx)$ is then evaluated and $\mathcal{GP}$ is updated with the result. Candidates are iteratively selected and evaluated until a budget constraint has been met.

In this work, we make use of the \texttt{BoTorch} optimization library \cite{balandat2020botorch} and use BO to improve the hyperparameters $\vtheta$ of the previously introduced structured control policies.
This is achieved by setting $f{=}\E_\vg\left[\mathcal{R}(\vtheta,\vg) \right]$, where $\vg$ represents the respective goal pose for the experiment and $ \mathcal{R}$ is the objective function used in the RRC.
The expectation over $\mathcal{R}$ is approximated by averaging over $N$ experiments, either in simulation or on the real platform.
Lastly, the uncertainty of the Gaussian process model is estimated through Mat\'ern $5/2$ kernels and the best candidate hyperparameters are obtained by maximizing the expected improvement acquisition function after fitting the model to the collected data.

\textbf{Residual Policy Learning}
In RL~\cite{sutton18}, an agent interacts with a MDP defined by the tuple $\mathcal{M}=\{\mathcal{S}, \mathcal{A}, \mathcal{T}, r, \gamma\}$. At any point in time, the agent is in a state $\vs_t \in \mathcal{S}$ and produces an
action $\va_t \in \mathcal{A}$. The MDP then transtions to a new state $\vs_{t+1} \in \mathcal{S}$ and the agent recevies a reward $r(\vs_t, \va_t)$. The goal of the agent is to select a policy $\pi: \mathcal{S} \rightarrow \mathcal{A}$ mapping states to actions that maximize the discounted sum of rewards: $\pi^*{=}\textstyle\argmax_\pi \; \E_\pi\left[\sum_{t=0}^T \gamma^t r(\vs_t, \va_t) \right]$.

In robotics, it is often easy to design a controller that obtains reasonable performance on the desired task. \Gls{rpl}~\cite{Silver2018ResidualPL, johannink2019residual} aims to leverage an existing controller $\pi_0$ and use RL to learn corrective or \textit{residual} actions to that controller. The learned policy acts by adding an action $\va^r_t \sim \pi(\vs_t, \va^0_t)$ to the action $\va^0_t \sim \pi_0(\vs_t)$ provided by the controller. 
From the agent's perspective, this is a standard RL setting with the state space and transition dynamics augmented by $\pi_0$: $\mathcal{M} = (\mathcal{S} \times \mathcal{A}, \mathcal{A}, \mathcal{T}', \mathcal{r}, \gamma)$, where $\mathcal{T}'([\vs_t,\va_t^0], \va^r_t, [\vs_{t+1}, \va_{t+1}^0]) = \mathcal{T}(\vs_t, \va^0_t + \va^r_t, \vs_{t+1}) \mathbb{P}(\va^0_{t+1} | \pi_0(\vs_{t+1}))$.
\Gls{rpl} benefits from the inductive bias of the base controller $\pi_0$, which can significantly improve exploration and sample efficiency.

In this work, we learn residual controllers on top of the three structured approaches defined above. To do this, we use soft actor-critic (SAC)~\cite{haarnoja2018soft}, a robust RL algorithm for control in continuous action spaces based on the maximum entropy RL framework.
SAC has been successfully used to train complex controllers in robotics~\cite{haarnoja2018learning}.
 
\section{EXPERIMENTS}

To provide a thorough benchmark of the above methods, we perform a series of detailed experiments and ablations in which we test the contribution of different components on L3 and L4 of the \gls{rrc}.

\textbf{Experiment Setup} In our experiments, we will be comparing different combinations of grasp strategies, controllers, and optimization schemes in a PyBullet \cite{pybullet} simulation environment and on the TriFinger platform. For each combination, we report the reward and final pose error averaged over several trials, as well as the fraction of trials the object is dropped. In simulation, we provide each method with the same initial object poses and goal poses. On the TriFinger platform, we cannot directly control the initial pose of the object, so we first move it to the center of the workspace and assign the goal pose as a relative transformation of the initial pose. All methods are tested with the same set of relative goal transformations. To isolate the performance of the grasp strategies and controllers, we initialize the experiments so that no cube alignment primitives are required to solve the task.

\textbf{Mix and Match} The choice of grasp heuristic and control strategy are both crucial to success, but it is hard to know how much each component contributes individually.
To test each piece in isolation, we ``mix and match" the three grasp heuristics with the three structured controllers and report the performance of all nine combinations. Each combination is tested on L4 and the results are averaged over $15$ trials.
We note that the speed and accuracy of the initial cube alignments had a large impact on the competition reward. 
To account for this and to test the robustness of these approaches to different alignment errors, we evaluate each method with three different initial orientation errors $\delta\theta \in \{10,25,35\}$ degrees. When the MP controller is paired with a grasp strategy other than PG, a motion plan is generated using that grasp.

\textbf{Bayesian Optimization} All of our approaches are highly structured and rely on a few hyperparameters. 
We investigate whether \gls{bo} can improve the performance of the controllers by optimizing them for both L3 and L4.
For all experiments, we initialize the optimization with four randomly sampled initial sets of parameters and run $50$ optimization iterations. We do not explicitly exploit any information from our manually tuned values. Instead, the user only has to specify intervals. 
This sample-efficient optimization process only takes about $12$ hours to complete on the real system. After the optimized parameters are found, each controller is tested over $20$ trials on the TriFinger platform with both the manually tuned parameters and the \gls{bo} parameters.
For our proposed approaches, we optimize the following values: CIC: gains and reference position  $\vx_r$; CPC: gains, including values for the exponential gain scheduling; and MP: hyperparameters that control the speed of movement to the goal location on the planned path.  

\textbf{Residual Policy Learning} In these experiments, we investigate to what extent \gls{rpl} can be used to improve the performance of our three controllers. To do this, we train a neural network policy to produce joint torques in the range $[-0.05, \, 0.05]$ (the maximum allowed joint torque on the system is $0.397$), which are then added to the actions of the base controller. The policies are trained on L3 in simulation for 1M timesteps using SAC \cite{haarnoja2018soft}. The reward function consists of a combination of the competition reward as well as terms for action regularization, maximizing tip force sensor readings, and maintaining a grasp of the object. The policy architecture is as follows: The observation and base action are separately embedded into a $64$-dimensional space. These embeddings are then concatenated and passed to a three-layer feed-forward network that outputs a Gaussian distribution over torque actions. Actions are sampled from this distribution, squashed using a $\tanh$ function, and then scaled to fit in the $[-0.05, \, 0.05]$ range. We evaluate MP-PG, CPC-TG, and CIC-CG in simulation over $20$ trials for L3. We then test their ability to transfer to the real system with another $20$ trials.

\section{RESULTS}
\begin{table*}[t]
    \vspace{5pt}
    \centering
    \begin{tabular}{ll|ccc|ccc|ccc}
\toprule
   &    & \multicolumn{3}{c|}{$\delta\theta = 10$\,deg} & \multicolumn{3}{c|}{$\delta\theta = 25$\,deg} & \multicolumn{3}{c}{$\delta\theta = 35$\,deg} \\
   &             &     Ori. Err.    &    Pos. Err. & Drop & Ori. Err. & Pos. Err.  & Drop & Ori. Err. &   Pos. Err. & Drop \\
Controller & Grasp &       [deg]       &       [cm]   & [\%] &  [deg]    &  [cm]      & [\%]  &  [deg]   &       [cm]  & [\%]         \\
\midrule
\multirow{3}{*}{CIC} & CG &  $ 55.9 \pm 27.37 $ &  $ 0.3 \pm 0.14 $ &   $ \hphantom{0}0.0 $ &  $ \mathbf{47.0 \pm 29.64} $ &  $ \mathbf{0.8 \pm 1.12} $ &   $ \hphantom{0}\mathbf{0.0} $ &  $ 58.2 \pm 20.95 $ &  $ 0.4 \pm 0.16 $ &   $ \hphantom{0}6.7 $ \\
   & PG &  $ 51.3 \pm 26.72 $ &  $ 0.4 \pm 0.60 $ &   $ \hphantom{0}0.0 $ &  $ 53.6 \pm 27.26 $ &  $ 0.7 \pm 0.96 $ &   $ \hphantom{0}0.0 $ &  $ \mathbf{62.4 \pm 31.69} $ &  $ \mathbf{0.4 \pm 0.24} $ &   $ \hphantom{0}\mathbf{0.0} $ \\
   & TG &  $ \mathbf{30.3 \pm 12.83} $ &  $ \mathbf{0.4 \pm 0.22} $ &   $ \hphantom{0}\mathbf{0.0} $ &  $ 39.4 \pm 20.38 $ &  $ 0.4 \pm 0.17 $ &  $ 13.3 $ &  $ 30.7 \pm 17.37 $ &  $ 0.4 \pm 0.21 $ &   $ \hphantom{0}6.7 $ \\
\cline{1-11}
\multirow{3}{*}{CPC} & CG &  $ 27.6 \pm 36.74 $ &  $ 3.0 \pm 4.61 $ &  $ 26.7 $ &    $ \hphantom{0}8.6 \pm \hphantom{0}5.24 $ &  $ 1.5 \pm 3.16 $ &  $ 33.3 $ &  $ 17.2 \pm 15.38 $ &  $ 3.3 \pm 5.26 $ &  $ 60.0 $ \\
   & PG &  $ 14.0 \pm 16.98 $ &  $ 2.6 \pm 4.65 $ &  $ 26.7 $ &  $ 36.2 \pm 39.84 $ &  $ 4.6 \pm 4.49 $ &  $ 33.3 $ &  $ \mathbf{35.4 \pm 45.88} $ &  $ \mathbf{6.3 \pm 6.14} $ &  $ \mathbf{13.3} $ \\
   & TG &    $ \mathbf{\hphantom{0}5.0 \pm \hphantom{0}3.19} $ &  $ \mathbf{0.5 \pm 0.26} $ &   $ \hphantom{0}\mathbf{6.7} $ &  $ \mathbf{20.3 \pm 48.96} $ &  $ \mathbf{1.5 \pm 3.26} $ &  $ \mathbf{26.7} $ &    $ 5.0 \pm 2.99 $ &  $ 0.4 \pm 0.19 $ &  $ 33.3 $ \\
\cline{1-11}
\multirow{3}{*}{MP} & CG &  $ 28.7 \pm 10.86 $ &  $ 0.4 \pm 0.43 $ &   $ \hphantom{0}0.0 $ &  $ 29.0 \pm 13.95 $ &  $ 0.5 \pm 0.47 $ &   $ \hphantom{0}0.0 $ &  $ 43.6 \pm 11.49 $ &  $ 0.6 \pm 0.61 $ &   $ \hphantom{0}0.0 $ \\
   & PG &  $ 30.5 \pm 14.56 $ &  $ 0.5 \pm 0.46 $ &   $ \hphantom{0}0.0 $ &   $ \mathbf{28.0 \pm \hphantom{0}8.70} $ &  $ \mathbf{0.7 \pm 0.60} $ &   $ \hphantom{0}\mathbf{0.0} $ &  $ \mathbf{30.5 \pm 10.85} $ &  $ \mathbf{0.6 \pm 0.80} $ &   $ \hphantom{0}\mathbf{0.0} $ \\
   & TG &  $ \mathbf{26.7 \pm 11.24} $ &  $ \mathbf{0.4 \pm 0.51} $ &   $ \hphantom{0}\mathbf{0.0} $ &  $ 32.8 \pm 15.22 $ &  $ 0.6 \pm 0.42 $ &   $ \hphantom{0}0.0 $ &  $ 37.1 \pm 13.31 $ &  $ 0.6 \pm 0.64 $ &   $ \hphantom{0}0.0 $ \\
\bottomrule
\noalign{\vskip 2mm}  
\end{tabular}
    \caption{Mix and match experiment on the real platform.
    $\delta\theta$ denotes the error between the goal and initial cube orientation, 
    \textit{Drop} is the fraction of episodes that the cube is dropped.
    \textit{Pos. Err.} and \textit{Ori. Err.} are the position and orientation errors at the end of the episodes.
    Each value is calculated from 15 episodes. Using MP provides the best trade-off between accuracy and reliability. Among the grasp strategies, TG yields best performance while the advantages of the PG become apparent in the most difficult setting (rightmost column) by improving the drop rate for all controllers.
    }
    \vspace{-.5cm}
    \label{tab:mix-and-match-real}
\end{table*}
\begin{table*}[t]
\centering
\begin{tabular}{l|ccc|ccc}
\toprule
\multicolumn{1}{l}{} & \multicolumn{3}{c}{L3 Real System} & \multicolumn{3}{c}{L4 Real System} \\
Control Policy & $\mathcal{R}$ & Pos.\ Err. {[}cm{]} & Ori.\ Err. {[}deg{]} & $\mathcal{R}$ & Pos.\ Err. {[}cm{]} & Ori.\ Err. {[}deg{]} \\
\midrule
CIC-CG         & $-7818.9 \pm 3332.2$  & $4.07 \pm 2.49$    & N/A       & $-6998.1 \pm 1840.5$  & $0.86 \pm 1.63$    & $42.18 \pm 27.40$        \\ 
CIC-CG w BO    & $\mathbf{-5613.6 \pm 2643.7}$  & $\mathbf{2.15 \pm 2.02}$    & N/A       & $-7534.4 \pm 2976.9$  & $1.17 \pm 2.55$    & $48.59 \pm 37.20$        \\ 
\midrule
CPC-TG         & $-2912.2 \pm 1738.0 $ & $1.43 \pm 3.51$    & N/A       & $-6150.8 \pm 2754.7 $ & $2.53 \pm 4.13$    & $27.06 \pm 48.68$        \\
CPC-TG w BO    & $\mathbf{-2130.5 \pm 1149.5}$  & $\mathbf{0.48 \pm 1.04}$     & N/A     & $\mathbf{-5046.5 \pm 1664.7}$  & $\mathbf{1.05\pm 1.60}$     & $\mathbf{13.95 \pm 24.76}$        \\ 
\midrule
MP-PG          & $-6267.3 \pm 4363.8 $ & $2.15 \pm 4.41$    & N/A       & $-7105.0 \pm 2109.1$  & $0.64 \pm 0.11$    & $25.62 \pm 15.04$        \\
MP-PG w BO     & $\mathbf{-4510.4 \pm 1412.4}$  & $\mathbf{0.53 \pm 0.71}$    & N/A      & $-7239.2 \pm 2257.7$  & $0.45 \pm 0.66$    & $22.39 \pm 11.90$        \\ 
\bottomrule
\noalign{\vskip 2mm}  
\end{tabular}
    \caption{Comparison of the manually chosen parameters with those identified using \gls{bo} on the real system for the L3 (left) and L4 (right) experiments. For L3, the newly obtained parameters result in significant improvements. For L4, only CPC-TG can be improved using \gls{bo}.}
    \label{tab:bo}
    \vspace{-.75cm}
\end{table*}
In this section, we describe the results of the experiments described above. In total, we conducted more than 20k experiments on the real system.

\textbf{Mix and Match}
Experiment results on the real platform are summarized in Table~\ref{tab:mix-and-match-real}.
\textit{Drop} refers to the percentage of the episodes that the cube is dropped.  \textit{Pos.} and \textit{Ori.\ Err.} are the position and orientation errors at the end of the episodes. The values are only calculated from those runs that do not drop the cube.
While the reactive control policies CIC and CPC run at \SI{100}{\hertz} and \SI{200}{\hertz}, respectively, the MP provides control signals with \SI{500}{\hertz} after precomputing the trajectory. 
As a common trend, orientation error increases as $\delta\theta$ gets bigger,
and except for CPC, position errors stay close to zero with low variance across different settings.

We find that CPC drops the cube more frequently than other approaches, and its performance varies significantly depending on the choice of grasp.
This is because it drives the fingertip positions to the goal pose without considering whether the grasp can be maintained.
Yet, in the cases in which it can retain its grasp, CPC achieves much lower orientation errors than CIC or MP.
While CIC is similar to CPC, it is more robust to drops at the expense of accuracy, because it explicitly considers the forces the fingertips need to apply to achieve a desired motion.
MP is the most robust against drops and grasp choices, because it attempts to move only to locations in which the selected grasp can be maintained.
This comes at the expense of orientation errors, since the planner may have only found a point near the goal for which the grasp is valid.

When comparing CG and TG, we find that TG performs better in terms of both orientation error and drop rate.
We hypothesize that this is the case because the triangle shape facilitates applying forces to the cube in all directions.
The benefits of the planned grasp (PG) become apparent when the initial orientation error is large, improving the drop rate across all three controllers.
This verifies the intuition for grasp planning that, when the required orientation change is large, it helps to carefully select a grasp that is feasible both at the initial pose and near the goal pose.

\textbf{Bayesian Optimization} 
Table~\ref{tab:bo} depicts the results from running \gls{bo} on the real system to optimize the controllers' hyperparameters.
As can be seen on the left-hand side, for the L3 experiments, the newly obtained hyperparameters significantly improve the policies' mean reward as well as the mean position errors.
Furthermore, although we only averaged across five rollouts during training, the improvements persist when evaluating the policies on $20$ newly sampled goal locations. Visually inspecting the rollouts, we conclude that running \gls{bo} results in higher gains such that the target locations are reached more quickly.

Repeating the same experiment for L4 yields the results presented on the right-hand side of Table~\ref{tab:bo}. As shown in the table, only the performance for the CPC control strategy can be improved significantly. 
Comparing the two sets of parameters, the \gls{bo} algorithm suggests using lower gain values. This results in a more stable and reliable control policy and increases performance. 
In general, even though we do not provide the manually obtained parameters as a prior to the \gls{bo} algorithm, for the other two approaches, the performance of the optimized parameters is still on par with the manually tuned parameters. We reason that for the MP approach, the two hyperparameters might not provide enough flexibility for further improvements, while the CIC controller might have already reached its performance limits. 
The results indicate that \gls{bo} is an effective tool to optimize and obtain performant hyperparameters for our approaches and mitigates the need for tedious manual tuning. 
\begin{table*}[t]
\vspace{5pt}
\centering
\begin{tabular}{l|ccc|ccc}
\toprule
\multicolumn{1}{l}{} & \multicolumn{3}{c}{Simulation} & \multicolumn{3}{c}{Transfer to the Real System} \\
Control Policy & $\mathcal{R}$ & Pos.\ Err. {[}cm{]} & Drop {[}\%{]} & $\mathcal{R}$ & Pos.\ Err. {[}cm{]} & Drop {[}\%{]} \\
\midrule
CIC-CG         & $\mathbf{-112 \pm 32.2}$  & $\mathbf{0.55 \pm 0.28}$    & $\mathbf{0.0}$       & $-4410 \pm 3304$  & $1.60 \pm 1.74$    & $10.0$  \\ 
CIC-CG w RPL    & $-570 \pm 350.6$  & $3.52 \pm 3.91$    & $10.0$      & $-4593 \pm 3964$  & $1.54 \pm 1.53$    & $10.0$      \\ 
\midrule
CPC-TG         & $-70.0 \pm 9.42$ & $0.25 \pm 0.15$    & $0.0$      & $\mathbf{-2150 \pm 860}$  & $\mathbf{0.50 \pm 0.41}$    & $0.0$    \\ 
CPC-TG w RPL    & $-74.4 \pm 4.97$  & $0.18 \pm 0.12$     & $0.0$     & $-2847 \pm 1512$  & $1.46 \pm 3.16$    & $0.0$      \\ 
\midrule
MP-PG          & $-122 \pm 58.2$  & $0.35 \pm 0.43$    & $0.0$     & $-8638 \pm 6197$  & $8.47 \pm 8.65$    & $40.0$    \\ 
MP-PG w RPL     & $\mathbf{-85.8 \pm 22.6}$  & $\mathbf{0.22 \pm 0.18}$    & $0.0$      & $\mathbf{-4573 \pm 1547}$  & $\mathbf{1.74 \pm 3.10}$    & $\mathbf{0.0}$     \\ 
\bottomrule
\noalign{\vskip 2mm}  
\end{tabular}
    \caption{The results of combining our controllers with RPL. Reward and final pose errors are shown for L3 in simulation and transferring the learned policies to the real system.
    Surprisingly, we find that for all controllers, the learned policies transfer directly to the real system, yet, only improving the MP controller.}
    \label{tab:rl}
    \vspace{-.85cm}
\end{table*}

\textbf{Residual Policy Learning} Table~\ref{tab:rl} shows the performance of our control strategies with and without \gls{rpl}. 
We find that the effectiveness of \gls{rpl} is limited to improving only the MP controller.
When inspecting the results we find that residual control is able to help the MP policy to maintain a tight grasp on the cube, preventing catastrophic errors such as dropping the cube and triggering another round of planning. 
Surprisingly, the learned policies are able to transfer to the real system without any additional finetuning or algorithms such as domain randomization~\cite{tobin2017domain}. For the MP controller, we find improved grasp robustness and a reduction in the drop rate. Additionally, for CPC and CIC, it is surprising that performance was not more adversely affected given the lack of improvement in simulation.
We hypothesize that this is the case for two reasons. First, the torque limits on the residual controller are small and may not be able to cause a collapse in performance.
Second, the base controllers provide increasing commands as errors are made that work to keep the combined controller close to the training data distribution. For example, the MP controller has a predefined path and the PD controller that follows that path provides increasing commands as deviations occur.

From these experiments, we conclude that \gls{rpl} may be effective when small changes can be made, such as helping to maintain contact forces, in order to improve the robustness of a controller, and that transferring a residual controller from simulation is much easier than transferring a pure RL policy.

\textbf{Challenge Retrospective}
Overall, the newly obtained results (e.g.,  Table \ref{tab:bo}) differ only slightly from to the scores reported during the competition (Table \ref{tab:challenge_results_phase2}).
CPC-TG yields the best results on L3, but the team using MP-PG was able to win, in part because they implemented more reliable primitives for cube alignment. Through exploiting insights from this approach, we assume that CPC-TG could perform similarly. Nevertheless, the robustness of MP-PG might still outweigh the gains on the successful runs of the reactive policies, especially with increasing task difficulty. 
During the third phase, all methods were also evaluated with a small cuboid and are thus not limited to a particular object type, as long as the grasping strategies are adapted accordingly. Yet, smaller objects make the problem of vision-based state estimation considerably more difficult, resulting in even further advantages of the MP-PG approach.
Looking broadly at the RRC and our results, it may be worth adjusting the RRC protocol to also provide an evaluation of the individual components (e.g., grasp planning) to yield more informative results. 
\section{CONCLUSION AND OUTLOOK}

In this work, we present three different approaches to solving the tasks from the \gls{rrc}.
We perform extensive experiments in simulation and on the real platform to compare and benchmark the methods.

We find that using motion planning provides the best trade-off between accuracy and reliability.
Compared to motion planning, the two reactive approaches vary in both reliability and accuracy.
Concerning grasp selection, the results show that using the triangle grasp yields the best performance.

We further show the effectiveness of running Bayesian optimization for hyperparameter tuning.
Especially for L3, the performance can be increased significantly across all approaches.
Augmenting the structured methods with a learned residual control policy can improve performance when small changes to a controller are beneficial.
Surprisingly, we also find that transferring the learned residual controllers required no finetuning on the real system or other techniques to cross the sim-to-real gap, although applying those techniques is likely to be beneficial.

We hope that our work serves as a benchmark for future competitions and dexterous manipulation research using the TriFinger platform.

\addtolength{\textheight}{-0cm}

\small
\bibliographystyle{IEEEtran}
\bibliography{references.bib}

\end{document}